\documentclass{article}

\usepackage[T1]{fontenc}
\usepackage{hyperref}
\usepackage{microtype}

\usepackage{xcolor}
\usepackage[nonatbib,final]{corl_2021}

\usepackage[center]{Styles/VABcommands}
\usepackage{Styles/VABenvironments}
\usepackage{Styles/VABcitations}
\usepackage{amsthm}
\usepackage{mathtools}
\usepackage{thmtools}
\usepackage{thm-restate}
\let\bm\boldsymbol
\usepackage[font=small,labelfont=small]{caption}
\usepackage{xurl}

\newcommand{\ty}[1]{{\scriptscriptstyle{\c{#1}}}}
\newcommand{\trsp}{\mathsf{T}}

\bibliography{references.bib}

\title{Geometry-aware Bayesian Optimization in Robotics using Riemannian Matérn Kernels}

\author{
  Noémie~Jaquier\textsuperscript{\ensuremath{*1}}
  \quad 
  Viacheslav Borovitskiy\textsuperscript{\ensuremath{*2,6}}
  \quad 
  Andrei Smolensky\textsuperscript{\ensuremath{2}} 
  \\ \bfseries
  Alexander Terenin\textsuperscript{\ensuremath{3,4}}
  \quad 
  Tamim Asfour\textsuperscript{\ensuremath{1}}
  \quad 
  Leonel Rozo\textsuperscript{\ensuremath{5}} 
  \\
  \textsuperscript{\ensuremath{1}}Karlsruhe Institute of Technology
  \quad 
  \textsuperscript{\ensuremath{2}}St. Petersburg State University 
  \\ 
  \textsuperscript{\ensuremath{3}}University of Cambridge
  \,
  \textsuperscript{\ensuremath{4}}Imperial College London
  \,
  \textsuperscript{\ensuremath{5}}Bosch Center for Artificial Intelligence
  \\
  \textsuperscript{\ensuremath{6}}St. Petersburg Department of Steklov Mathematical Institute of Russian Academy of Sciences
}

\begin{document}
\maketitle

\begin{abstract}
Bayesian optimization is a data-efficient technique which can be used for control parameter tuning, parametric policy adaptation, and structure design in robotics.
Many of these problems require optimization of functions defined on non-Euclidean domains like spheres, rotation groups, or spaces of positive-definite matrices.
To do so, one must place a Gaussian process prior, or equivalently define a kernel, on the space of interest.
Effective kernels typically reflect the geometry of the spaces they are defined on, but designing them is generally non-trivial.
Recent work on the Riemannian Matérn kernels, based on stochastic partial differential equations and spectral theory of the Laplace--Beltrami operator, offers promising avenues towards constructing such geometry-aware kernels.
In this paper, we study techniques for implementing these kernels on manifolds of interest in robotics, demonstrate their performance on a set of artificial benchmark functions, and illustrate geometry-aware Bayesian optimization for a variety of robotic applications, covering orientation control, manipulability optimization, and  motion planning, while showing its improved performance.
\end{abstract}

\keywords{Bayesian optimization, Mat\'ern kernels, Riemannian manifolds}

{
\renewcommand\thefootnote{}\footnotetext[0]{\textsuperscript{\ensuremath{*}}Equal contribution. Correspondence to: \href{mailto:noemie.jaquier@kit.edu}{\textsf{noemie.jaquier@kit.edu}}, \href{mailto:viacheslav.borovitskiy@gmail.com}{\textsf{viacheslav.borovitskiy@gmail.com}}. Code at \url{https://github.com/NoemieJaquier/MaternGaBO} and video at \url{https://youtu.be/6awfFRqP7wA}.}
}

\section{Introduction}
\label{sec:Introduction}

Fast and data-efficient adaptation is a key challenge in a large range of robotics applications, owing to the need to handle sensor noise, model uncertainty, and generalize to unforeseen settings.
In this context, Bayesian optimization~\cite{snoek2012} is a technique of growing interest in robotics due to its ability to efficiently solve challenging optimization problems, including controller tuning~\cite{Marco16:LQRbayesOpt, Antonova17:DeepKernelsBO}, policy adaptation~\cite{Akrour2017,Froelich2021}, and robot design~\cite{Saar2018}.
This makes it an attractive tool for settings where data-efficiency is of key interest, and large-scale methods such as deep reinforcement learning are not presently viable.
Bayesian optimization algorithms are generally powered by probabilistic models based on Gaussian processes that rely on inductive biases encoded by kernels.
Improving performance and scalability of such models via a number of different strategies has therefore been a central goal pursued by the Bayesian optimization and Gaussian process literatures.

Inductive bias in robotics can be introduced by exploiting domain knowledge about the task at hand. 
For example, task-specific kernels were proposed by~\textcite{Antonova17:DeepKernelsBO} by learning a distance metric via simulated bipedal locomotion patterns to optimize gait controllers, and by \textcite{Rai2018} using gait feature transformations to design kernels to optimize locomotion controllers.
More generic kernels were introduced by~\textcite{Marco16:LQRbayesOpt,Marco17:LQRkernel,Wilson14:BO_TrajectoryKernel} for policy optimization by incorporating linear-quadratic control structures and trajectory information into the kernel design.
While these kernels can be exploited for robot control or policy search with many different tasks and systems, their use is specific to the problems they were designed for.

\begin{figure}
\centering
\hfill
\includegraphics[scale=0.175]{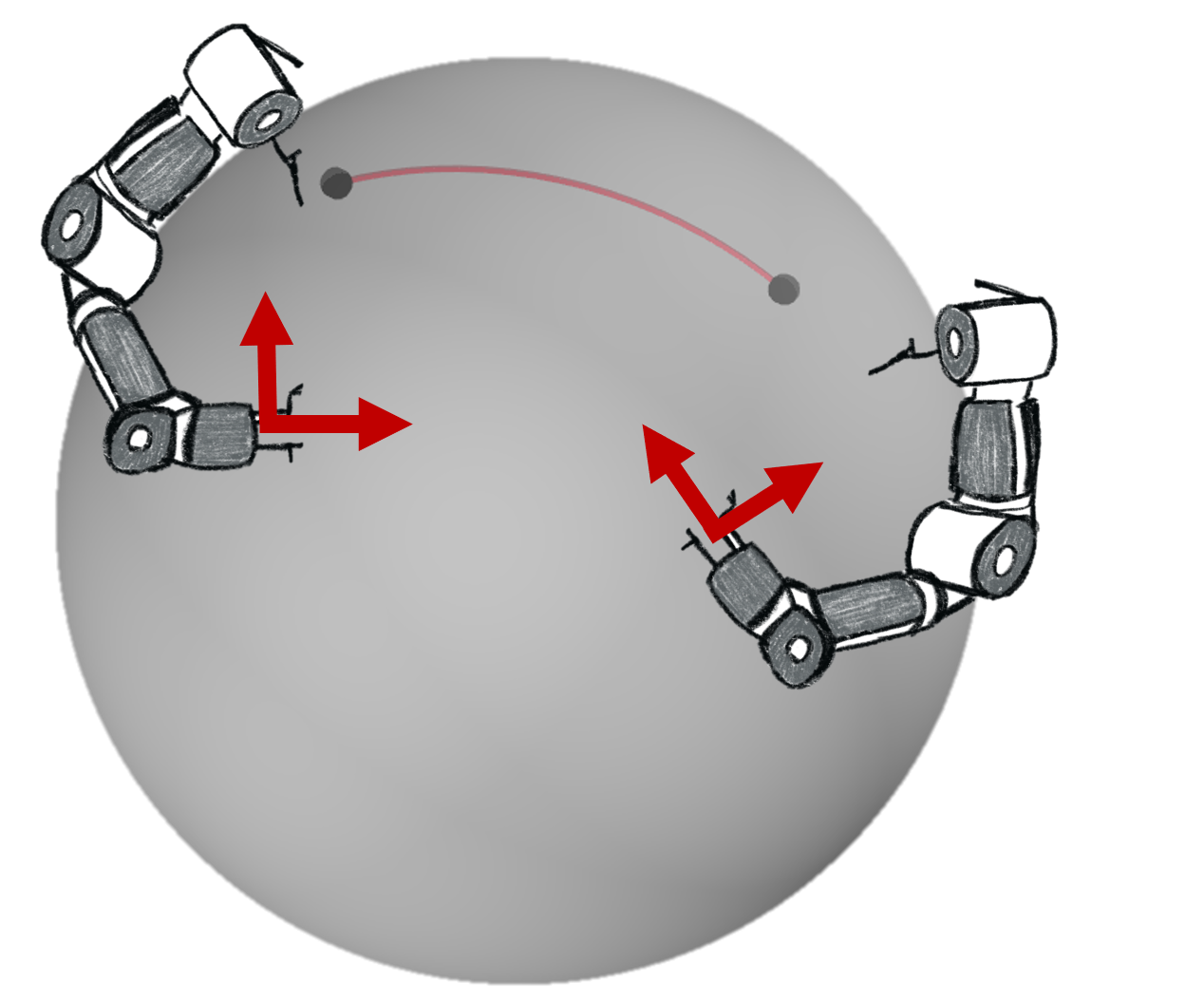}
\hfill
\includegraphics[scale=0.175]{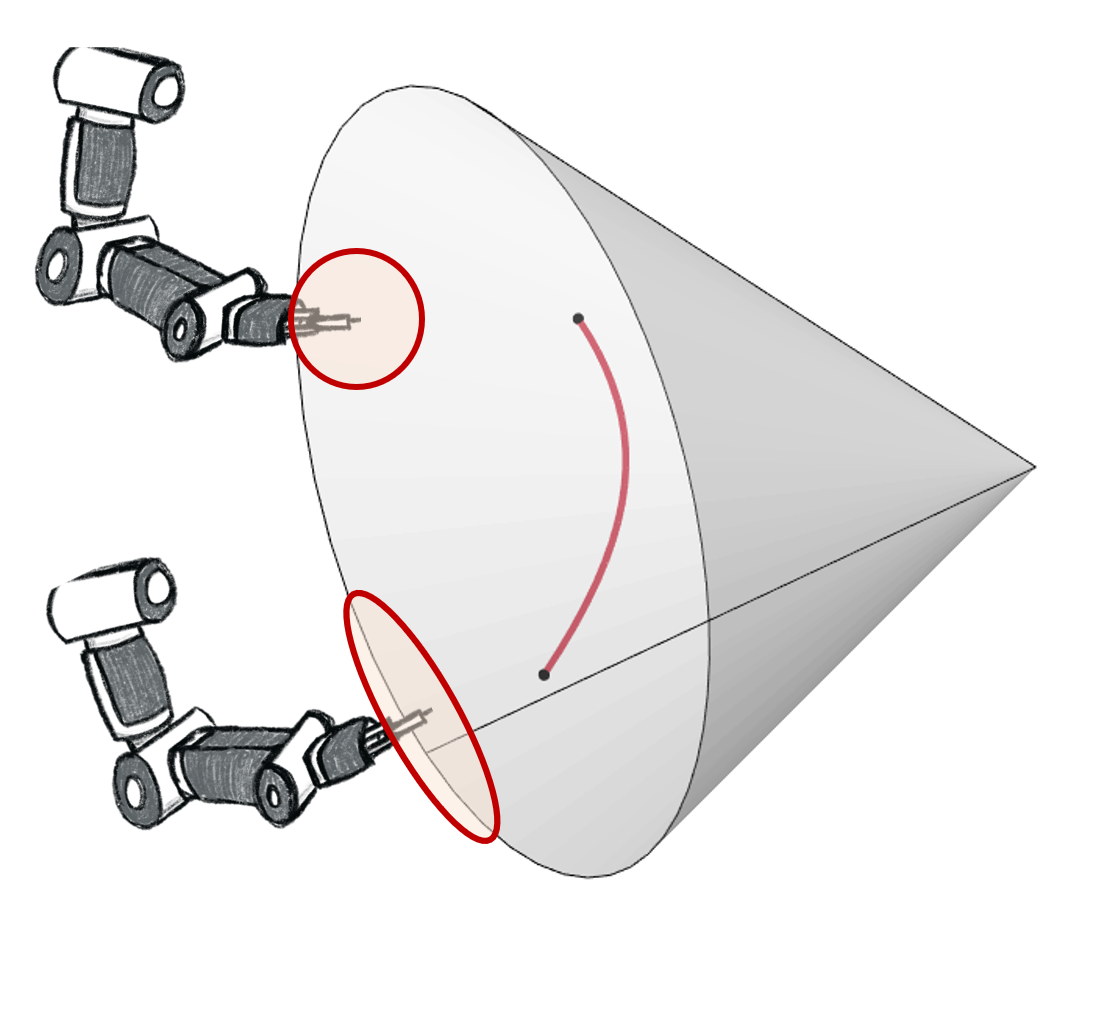}
\hfill
\includegraphics[scale=0.175]{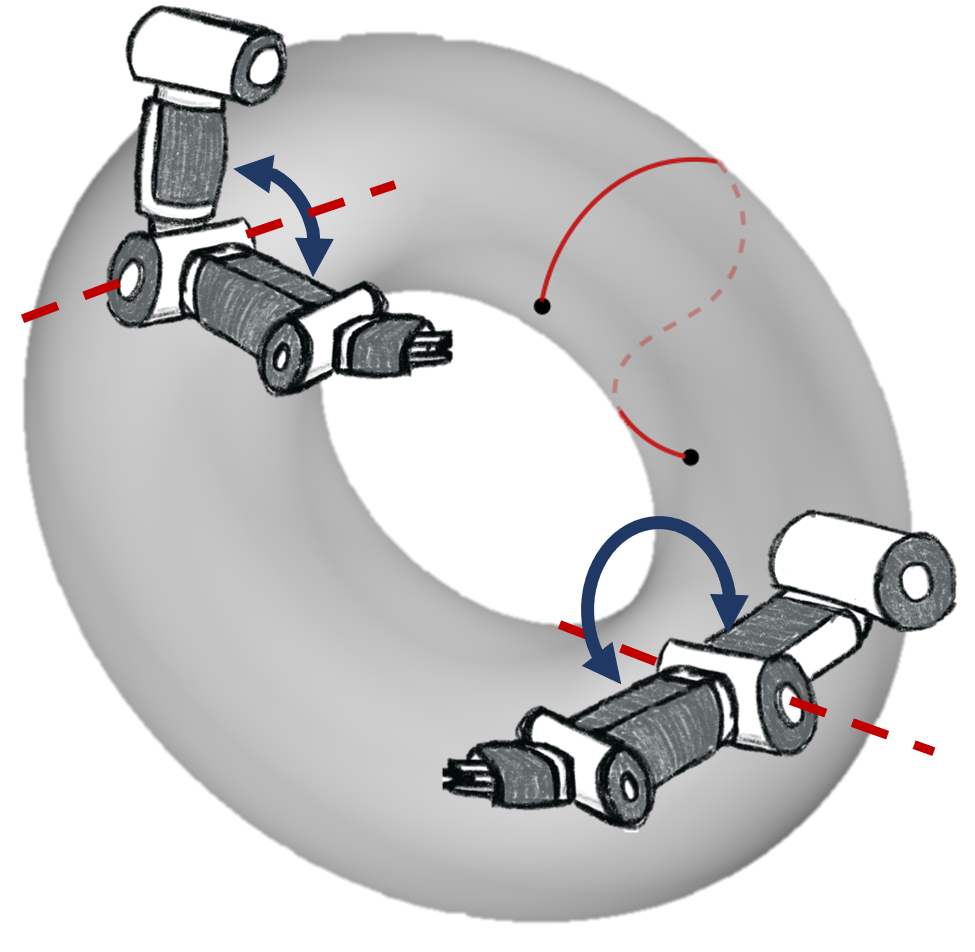}
\hfill
\includegraphics[scale=0.175]{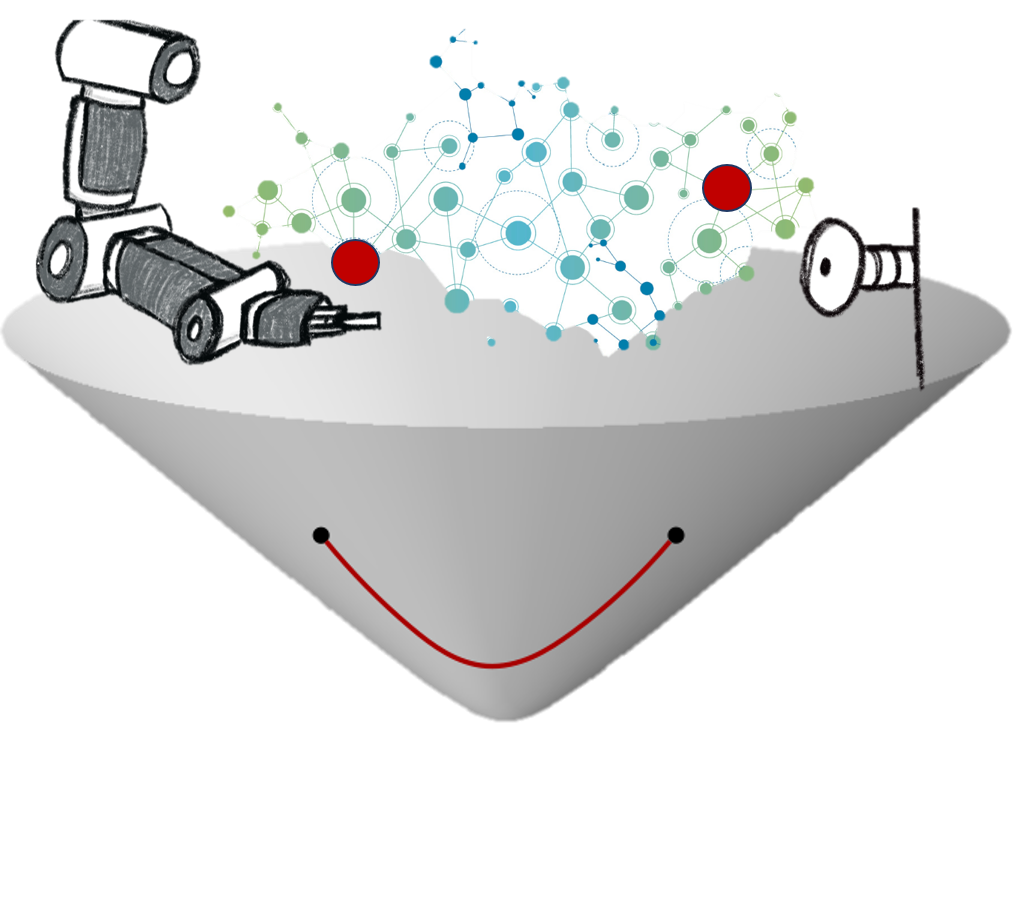}
\hfill
\caption{Illustration of different Riemannian manifolds of interest in robotics: a unit sphere $\mathbb{S}^d$, symmetric positive definite matrices $\c{S}_{++}^d$, a torus $\T^d$, and the Lorentz model of hyperbolic geometry in $\c{H}^2$ (from left to right). Red solid lines depict geodesics between two (black) points on the manifold.}
\vspace{-0.25cm}
\label{fig:manifolds_illustration}
\end{figure}

A more general approach is to carefully select the \emph{domain} of the function being optimized, thus incorporating information about the geometry of the search space into the optimization algorithm.
Many quantities of interest in robotics carry such geometric information: three-dimensional rotations can be viewed as elements of the Lie group $\operatorname{SO}(3)$~\cite{Gu1988} or the sphere $\mathbb{S}^3$~\cite{Wen1991}, control gains, inertia and manipulability ellipsoids lie in the manifold of symmetric-positive-definite matrices~$\c{S}_{\ty{++}}^d$, while the joint configuration of a $d$-degree-of-freedom robot with revolute joints may be viewed as a point on a torus $\T^d$, i.e. the product of $d$ circles \cite{Kavraki2016}: see Figure \ref{fig:manifolds_illustration} for illustrations.
Incorporating this geometric structure into Bayesian optimization has recently been shown to improve its performance on a number of tasks~\cite{Jaquier19CoRLa}, despite the use of a naïve geometrical approximation in the kernel design.

Geometry-aware Bayesian optimization is therefore a promising emerging tool for general use in robotics.
One of the main difficulties in developing this tool is that defining geometry-aware Gaussian process models is technically non-trivial: classic Euclidean methods ignore the parameter space geometry, and naïve geometric approximations do not capture the manifold's geometric structure.
Fortunately, mathematical theory and techniques for building the necessary kernels are beginning to emerge \cite{lindgren2011,Borovitskiy_20NeurIPS}.
In particular, the work of~\textcite{Borovitskiy_20NeurIPS} on Riemannian Matérn kernels provides mathematically-sound tools for calculating geometry-aware kernels by building on stochastic partial differential equations and spectral theory of the Laplace--Beltrami operator.
This provides the key ingredients to deploy geometry-aware Bayesian optimization on practical robotics problems.

Building on the performance improvements in~\textcite{Jaquier19CoRLa} and the theoretically-grounded Matérn kernels proposed in~\textcite{Borovitskiy_20NeurIPS}, we present a more general kernel formulation for a wide range of Riemannian manifolds of interest in robotics.
Our approach uses the fact that a Matérn kernel can be formulated as the integral of a squared-exponential kernel: this idea enables us to define Riemannian Matérn kernel on non-compact manifolds, generalizing the work of~\textcite{Borovitskiy_20NeurIPS} which applies only to compact manifolds.
We introduce new Riemanian Matérn kernels for the special orthogonal group, manifold of symmetric positive definite matrices, and hyperbolic space, and study their use in robotic applications.
This more general formulation avoids the need to determine valid length scales as in prior work \cite{Jaquier19CoRLa}, provides theoretical guarantees of positive-definiteness, and makes kernel smoothness optimization possible. 
We benchmark these kernels on a set of artificial functions and on a number of robotics problems of interest to practitioners.

\section{Geometry-aware Bayesian Optimization}
\label{sec:Background}
Bayesian optimization \cite{Shahriari16:BayesOpt} is a class of sequential search algorithms for finding a global minimizer $\v{x}^* = \argmin_{\v{x}\in\c{X}} f(\v{x})$ of an unknown objective function $f$ defined over a domain of interest $\c{X}$.
The function $f$ is not observed directly: instead, at each step the algorithm must select a query point $\v{x}\in\c{X}$ at which to evaluate $f$.
Moreover, it is usually assumed that the algorithm only observes a noisy value $y\in\R$ whose expectation $\E(y) = f(\v{x})$ is the true function value at the chosen point.
Based on the full set of observed evaluations, the algorithm must select the next query point $\v{x}$.

The vast majority of Bayesian optimization algorithms employ Gaussian process models, which represent the known information about the objective function using a Gaussian process prior $f\sim\operatorname{GP}(\mu,k)$, with mean function $\mu : \c{X} \to \R$ and symmetric-positive-definite kernel $k : \c{X} \times \c{X} \to \R$. 
The Gaussian process is updated by conditioning on observed data $(\v{x}_i,y_i)_{i=1}^n$ using Bayes' Rule.
The next query point is selected by optimizing a carefully-constructed acquisition function defined with respect to the resulting posterior distribution.

In robotics, the domain of interest $\c{X}$ often possesses important non-Euclidean geometric structure.
For example, when tuning robot controllers, the variables of interest include end-effector poses lying on $\R^3 \times \mathbb{S}^3$ where $\mathbb{S}^d$ denotes the $d$-dimensional unit sphere, robot joint postures defined on a torus~$\T^d = \mathbb{S}^1 \x \ldots \x \mathbb{S}^1$, as well as gain matrices and manipulability ellipsoids belonging to the space of symmetric-positive-definite matrices $\c{S}_{\ty{++}}^d$.
In addition, when adapting parametric robot policies, some of their parameters such as mean vectors or covariance matrices lie in Riemannian manifolds.
This inherent geometry is often overlooked in robotic applications~\cite{Marco16:LQRbayesOpt, Rozo19:ForceBO, Roveda2021}. 
In this paper, we demonstrate that incorporating the correct geometric structure of $\c{X}$ into the Gaussian process is an important potential avenue for improving performance of Bayesian optimization.

Performing geometry-aware Bayesian optimization on Riemannian manifolds requires one to address two main challenges.
Firstly, defining a valid Gaussian process prior is more difficult than in the Euclidean case.
Secondly, numerical techniques such as training and optimizing the acquisition functions may require special considerations.
We address these issues in the remainder of this section.

\subsection{Gaussian Processes on Riemannian Manifolds}

The key challenge in constructing Gaussian processes on non-Euclidean domains is in defining a valid kernel: seemingly obvious and innocuous-looking candidate expressions such as the squared exponential geodesic distance $\exp(-d_g(\v{x},\v{x}')^2/\kappa)$ are not valid kernels for all length scales $\kappa > 0$ simultaneously~\cite{feragen2015}.
In spite of this, they have been shown to be beneficial for practical robotics applications~\cite{Jaquier19CoRLa}: this motivates one to develop technically sound analogues.
To tackle these issues, recent works on non-Euclidean Gaussian processes, including~\textcite{lindgren2011} and \textcite{Borovitskiy_20NeurIPS,borovitskiy21}, have developed provably well-posed techniques based on stochastic partial differential equations and spectral theory of the Laplace--Beltrami operator.
We focus on the latter work as it does not require one to solve stochastic partial differential equations numerically.

The general theory of Matérn Gaussian processes on compact Riemannian manifolds from~\textcite{Borovitskiy_20NeurIPS} introduces expressions for the kernels of these processes in terms of the Laplace--Beltrami eigenpairs of their respective manifolds.
For more general settings including non-compact manifolds, such as the manifold of positive definite matrices and hyperbolic space, kernels can often be built by exploiting the connection between Riemannian squared exponential and heat kernels as fundamental solutions of heat equations.
These can be connected with Matérn kernels by viewing them as certain integrals of squared exponential kernels~\cite{Tronarp2018}.
We use this connection to introduce new Riemannian kernels on non-compact manifolds, and to further explore existing Riemannian kernels in Section~\ref{sec:GaBO}.

\subsection{Optimization of acquisition functions on manifolds}

To use these Gaussian processes in a Bayesian optimization system, one must further be able to determine where to evaluate the unknown function and collect additional data. To do so, Bayesian optimization minimizes an acquisition function, which resolves the \emph{explore-exploit tradeoff} \cite{Shahriari16:BayesOpt} inherent in the problem due to lack of complete information about $f$.
This is done, for instance, by prioritizing points which are known to perform well due to having high posterior mean, while also considering points whose performance is unknown due to high posterior variance.
In this way, the probabilistic uncertainty given by the posterior Gaussian process is used to resolve the tradeoff.

In the geometric setting, Bayesian optimization therefore requires minimization of real-valued functions on manifold domains as an intermediate step.
To do so, we leverage optimization algorithms on Riemannian manifolds~\cite{Absil07:book} to maximize the acquisition function in a geometry-aware manner.
These algorithms work with the Euclidean tangent space $\c{T}_{\v{x}} \c{X}$ linked to each point $\v{x}$ on the manifold~$\c{X}$, utilize gradients intrinsically defined as elements of the tangent space, and project the results of their computations back onto the manifold with the exponential map $\text{Exp}_{\v{x}}: \c{T}_{\v{x}} \c{X}\to \c{X}$.
These different operations are determined based on the Riemannian metric with which the manifold is endowed.  

In this paper, we employ trust region methods on Riemannian manifolds, as introduced by~\textcite{Absil07:TR}, to optimize the acquisition function at each iteration of Bayesian optimization.
The recursive process used in trust region methods on Riemannian manifolds involves the same steps as its Euclidean counterpart, namely: (i) optimization of a quadratic subproblem $m_k$ trusted locally in a region around the iterate, (ii) updating of the trust region parameters, typically the trust-region radius $\Delta_k$, (iii) updating the iterates, where a candidate is accepted or rejected as function of the quality of the model $m_k$.
For the cases where the space of interest is restricted to a subset of a Riemannian manifold, we employ a constrained trust region algorithm proposed by~\textcite{Jaquier20NeurIPS} that handles linear constraints.
This recipe applies to almost all acquisition functions, including stochastic ones via pathwise conditioning \cite{wilson20,wilson21}, and allows us to introduce physical or safety constraints that naturally arise when interacting with physical systems and are crucial for robotics applications. 

\section{Matérn Kernels on Riemannian Manifolds}
\label{sec:GaBO}

Training Gaussian processes, a crucial component of Bayesian optimization, requires us to be able to compute the kernel point-wise. 
This kernel encapsulates similarity information in the domain of interest: for domains with geometric structure, such as Riemannian manifolds, the kernel should reflect this structure in order to perform effectively.
We now introduce geometric kernels that capture the structure of a number of spaces of interest in robotics, and study techniques for computing them.

As aforementioned, we focus on Riemannian Matérn kernels.
These are based on the characterization of the widely used Euclidean Matérn Gaussian processes as solutions of a class of stochastic partial differential equations given by~\textcite{whittle1963}.
These equations naturally generalize to the Riemannian setting, and yield kernels which are well-defined for all Riemannian manifolds and hyperparameter values, unlike prior work \cite{Jaquier19CoRLa} where restricting length scale range is required to guarantee positive definiteness.
The main difficulty of this approach is that this definition is implicit and does not give practical expressions for computing the kernel without further considerations.
For compact Riemannian manifolds, \textcite{Borovitskiy_20NeurIPS} prove expressions for Matérn kernels of the form
\[
\label{eqn:mani-matern-formula}
k_{\nu, \kappa, \sigma^2}(\v{x}, \v{x}') &= \frac{\sigma^2}{C_\nu}\sum_{n=0}^\infty \Phi_\nu(\lambda_n) f_n(\v{x})f_n(\v{x}') 
&
\Phi_\nu(\lambda_n) &= \begin{cases}
\del{\frac{2 \nu}{\kappa^2} + \lambda_n}^{-\nu-\frac{d}{2}}
&
\nu < \infty,
\\
e^{-\frac{\kappa^2}{2} \lambda_n}
&
\nu = \infty,
\end{cases}
\]
where $(\lambda_n, f_n)$ are Laplace--Beltrami eigenpairs ($\lambda_n \geq 0$), $\kappa$ is the length scale, $\sigma^2$ is the variance, $\nu$ is the smoothness and the constant $C_{\nu}$ is chosen so that the resulting kernels have average variance across the domain equal to $\sigma^2$.
Note that despite notation $C_{\nu}$ depends on both $\nu$ and $\kappa$.
The limiting kernel as $\nu\to\infty$, which we denote by $k_{\infty, \kappa, \sigma^2}$, can be viewed as the Riemannian analog of the Euclidean squared exponential kernel.

In practice, the kernel~\eqref{eqn:mani-matern-formula} and its squared exponential limit are computed by calculating the eigenpairs either analytically or numerically, and then truncating the infinite sum.
Numerical computation of the eigenpairs amounts to approximating the manifold with a mesh and solving the eigenproblem using finite element methods \cite{lord14}.
Since this approach scales exponentially with manifold dimension, we do not focus on it in this work.
Fortunately, owing to their extensive use in physics and other areas, mathematical techniques for evaluating the eigenpairs and simplifying the related expressions are available for many manifolds of interest in robotics: examples of this can be seen in the sequel.
Although we only present the formulas for squared exponential (heat) kernels for the sake brevity, respective formulas for Matérn kernels hold as well.

The expression~\eqref{eqn:mani-matern-formula} does not apply for non-compact manifolds, although the implicit stochastic partial differential equation definition it is derived from is still valid.
In this case, we leverage representations of Matérn kernels as integrals of squared exponential kernels, also known as heat kernels in the mathematical literature.
In the Euclidean case, observe that
\[ \label{eqn:matern_integral_formula}
k_{\nu, \kappa, \sigma^2}(\v{x}, \v{x}')
=
C
\int_0^{\infty}
u^{\nu - 1}
e^{-\frac{2 \nu}{\kappa^2} u}
k_{\infty, \sqrt{2 u}, \sigma^2}(\v{x}, \v{x}')
\d u ,
\]
where $C$ is given in \eqref{eqn:matern_integral_formula_full}.
This can be obtained by writing the spectral measure of the Matérn kernel, the T distribution, as a gamma mixture of Gaussians.
A similar relationship also holds in the compact Riemannian case: see Appendix~\ref{appdx:theory} for details.
More generally, it can be considered as the \emph{definition} of a Matérn kernel in any setting where a suitable notion of a squared exponential or heat kernel is available (in this case we usually use $C = 1$), which is particularly useful in non-compact settings.
Moreover, as shown in Appendix~\ref{appdx:theory}, this kernel is positive definite whenever the respective heat kernel is positive definite. 
Therefore, it suffices to obtain expressions for the squared exponential kernel: analogous expressions for the Matérn kernel follow by integrating over length scales via~\eqref{eqn:matern_integral_formula}.
We thus focus on the heat kernel henceforth.\footnote{Note that it is usually beneficial to exploit explicit expressions for the Matérn kernels when they are available, e.g., \eqref{eqn:mani-matern-formula}, to alleviate additional numerical errors introduced by approximately computing the integral~\eqref{eqn:matern_integral_formula}.} 

These general expressions for Riemannian Matérn kernels allow us to build theoretically-sound kernels for the torus and the sphere. 
More importantly, we can introduce Riemannian formulations for kernels on non-compact manifolds such as the manifold of symmetric positive definite matrices, and $\c{H}^d$.
We now turn our attention to manifolds of key importance in robotics applications.

\paragraph{Torus.}
\label{sec:torus_kernel}
The torus $\T^d$ is defined as a product of circles $\T^d = \mathbb{S}^1 \x \ldots \x \mathbb{S}^1$, which naturally arises as, for example, the space of joint configuration of a $d$-degree-of-freedom robot.
Functions on a torus $f : \T^d \to \R$ can be viewed as $1$-periodic functions $f : \R^d \to \R$.
Using this identification, the Laplace--Beltrami eigenfunctions on $\T^d$ are exactly those eigenfunctions of the Euclidean Laplacian which are $1$-periodic: sines and cosines whose frequency is an integer multiple of $2 \pi$.
Substituting this into the general formula yields 
\[ \label{eqn:heat_torus}
k_{\infty, \kappa, \sigma^2}(\v{x},\v{x}') = \frac{\sigma^2}{C_\infty}
\sum_{\v{\tau} \in \Z^d}
e^{-2 \kappa^2 \pi^2 \norm{\v\tau}^2}
\cos(2 \pi \innerprod{\v{\tau}}{\v{x}-\v{x}'}) ,
\]
where the sum is over vectors of integers $\v{\tau}$, and $\v{x}, \v{x}' \in [0,1)^d$ are the representation of elements of $\T^d$ as vectors of angles divided by $2 \pi$, i.e., such that $\del[1]{e^{2 \pi i x_1}, \ldots, e^{2 \pi i x_d}}$ and $\del[1]{e^{2 \pi i x_1'}, \ldots, e^{2 \pi i x_d'}}$ are the corresponding elements of $\T^d$. 
Here, $C_\infty$ is the normalizing constant chosen so that $k(\v{x}, \v{x}) = \sigma^2$.
The key difference with \eqref{eqn:mani-matern-formula} is that the eigenpairs are now parameterized by vectors of integers, slightly simplifying the expression.
A simple proof of this expression is given in Appendix~\ref{appdx:theory}.

\paragraph{Sphere.}
\label{sec:sphere_kernel}
The sphere manifold occurs in a number of settings: in particular, unit quaternions commonly used in robotics to represent the orientation of rigid bodies are a common example of the three-dimensional sphere $\mathbb{S}^3$.
On the $d$-dimensional sphere $\mathbb{S}^d$ the squared exponential kernel is
\[
k_{\infty, \kappa, \sigma^2}(\v{x},\v{x}') = \frac{\sigma^2}{C_\infty} \sum_{n=0}^\infty c_{n,d}\,e^{-\frac{\kappa^2}{2} n (n+d-1)}\,\c{C}_n^{(d-1)/2} \del[1]{\cos\del[0]{d_g(\v{x},\v{x}')}}
,
\]
where $C_\infty$ is a normalizing constant which ensures $k_{\infty, \kappa, \sigma^2}(\v{x}, \v{x}) = \sigma^2$; $c_{n,d}$ are explicit constants, $\c{C}_n^{(\cdot)}$ are the Gegenbauer polynomials, and $d_g$ is the geodesic distance on the sphere~\cite{Borovitskiy_20NeurIPS}.
This kernel is obtained by noting that the eigenfunctions of the spherical Laplace--Beltrami operator are the spherical harmonics, sums of which may be reformulated through the Gegenbauer polynomials---the resulting algebra, along with combining repeated eigenvalues, leads to the desired expression. 
This expression is a significant simplification compared to the general case: note in particular that the kernel factorizes and can be written as a function only of the geodesic distance $d_g$.

\paragraph{Special orthogonal group.}
\label{sec:so_kernel}

The special orthogonal group $\operatorname{SO}(d)$ is a compact Riemannian manifold\footnote{$\operatorname{SO}(d)$ is a Lie group, meaning that it is a group endowed with a structure of differentiable manifold: this manifold admits a canonical Riemannian structure induced by the Killing form.} of rotations of $d$-dimensional Euclidean space, most often represented as the set of orthogonal $d\times d$ matrices of determinant $1$.
$\operatorname{SO}(3)$ is widely used to represent orientations in the $3$-dimensional physical space.
This manifold may likewise be useful as a building block for representing other geometric objects.
For instance, the product $\R^d \x \operatorname{SO}(d)$ may be used as a proxy to represent symmetric matrices via a non-unique decomposition $\m{A} = \m{U} \m{D} \m{U}^{\top}$, where $\m{D} = \operatorname{diag}\del{\lambda_1, \ldots, \lambda_d}$, $\lambda_i \in \R$ and $\m{U} \in \operatorname{SO}(d)$.
A kernel on $\R^d \x \operatorname{SO}(d)$ may be designed as a product of kernels on $\R^d$ and on $\operatorname{SO}(d)$.
By restricting the Euclidean part of the product kernel to $\R_{\geq 0}^d$ or $\R_{> 0}^d$, we may obtain a kernel over positive (resp. semi)definite matrices, albeit not a canonical Matérn kernel.

By leveraging the group structure of $\operatorname{SO}(d)$ it is possible to express the corresponding heat kernel in terms of \emph{group characters}.
Specifically, we have that
\[ \label{eqn:sod_heat}
k_{\infty, \kappa, \sigma^2}(\m{X}, \m{Y})
=
\frac{\sigma^2}{C_{\infty}}\sum_{\pi}
e^{-\frac{\kappa^2}{2} \lambda_{\pi}} d_{\pi} \chi_{\pi}(\m{X} \m{Y}^{-1}) ,
\]
where the sum is over a certain set of tuples of nonnegative integers $\pi$ called the \emph{highest weights} and $C_{\infty}$ is the normalizing constant ensuring that $k_{\infty, \kappa, \sigma^2}(\m{X}, \m{X}) = \sigma^2$: see Appendix \ref{appdx:theory}.
Here, $\lambda_{\pi}$ are Laplacian eigenvalues, $\chi_{\pi}$ are certain functions called \emph{characters} that depend only on eigenvalues of their argument, and $d_{\pi} = \chi_{\pi}(\m{I})$.
Partial sums of~\eqref{eqn:sod_heat} corresponding to the most significant eigenvalues may be computed in closed form by means of the Weyl character formula and Freudenthal's formula for the eigenvalues of the Casimir operator.
We~provide technical details concerning this formula and its practical implementation in Appendix~\ref{appdx:theory} and in the supplemented code respectively, while deferring a thorough discussion to future work.

\paragraph{Hyperbolic space.}
\label{sec:hyperbolic_kernel}

The hyperbolic space $\c{H}^d$ is the unique simply-connected complete $d$-dimensional Riemannian manifold with a constant negative sectional curvature $-1$.
An important property of this space is the exponential rate of growth of the volume of a ball with respect to its radius.
Because of this, it is often used to hold embeddings of hierarchical data, such as trees or graphs~\cite{Nickel2017,Chami2020}.
Although its potential to embed discrete data structures into a continuous space is well-known in the machine learning community, its application in robotics is presently scarce. 
In Section~\ref{sec:exp}, we provide a toy example featuring a simple path planning on a 2D grid to illustrate how the hyperbolic manifold can be exploited in robotics. 

Hyperbolic space is non-compact, thus the technique of~\textcite{Borovitskiy_20NeurIPS} is unable to represent this kernel.
Fortunately, explicit formulas for the corresponding heat kernel in arbitrary dimension $d$ are present in the literature.
Take $\v{x}, \v{y} \in \c{H}^d$ and denote by $\rho = \operatorname{dist}_{\c{H}^d}(\v{x}, \v{y})$ the geodesic distance between $\v{x}$ and $\v{y}$ in $\c{H}(d)$.
Then we have, as introduced in \textcite{grigoryan1998},
\[ \label{eqn:hyp_heat}
&\!\!\!\!\!k^{\c{H}(2)}_{\infty, \kappa, \sigma^2}(\v{x}, \v{y})
=
\frac{\sigma^2}{C_{\infty}}
\int_{\rho}^{\infty}
\frac{s \exp\del{-s^2/(2 \kappa^2)}}{\del{\cosh(s) - \cosh(\rho)}^{1/2}} \d s,
&
&k^{\c{H}(3)}_{\infty, \kappa, \sigma^2}(\v{x}, \v{y})
=
\frac{\sigma^2}{C_{\infty}}
\frac{\rho}{\sinh{\rho}} e^{-\frac{\rho^2}{2 \kappa^2}}
\]
for $d = 2$ and $d = 3$, and the following recurrence relation, known as Millson's formula, for $d > 3$:
\[
k^{\c{H}(d)}_{\infty, \kappa, \sigma^2}(\v{x}, \v{y})
=
-
\frac{\sigma^2}{C_{\infty} \sinh{\rho}}
\frac{\partial}{\partial \rho}
k^{\c{H}(d-2)}_{\infty, \kappa, \sigma^2}(\v{x}, \v{y}).
\]

\paragraph{Manifold of positive definite matrices.}
\label{sec:spd_kernel}

Positive definite matrices occur naturally in many physics related problems.
In robotics, they are used to represent the stiffness and manipulability ellipsoids~\cite{Ajoudani2017,Yoshikawa1985} or as control gain matrices~\cite{Marco16:LQRbayesOpt}.
These matrices strongly affect the performance of robot motion and force controllers, and they are variables of interest for task-compatible trajectory optimization. 

Symmetric positive definite (SPD) matrices may be endowed with different Riemannian structures.
For instance, $\m{A} \mapsto \exp(\m{A})$ and $\m{A} \mapsto \m{A} \m{A}^{\top}$ define one-to-one maps between symmetric matrices or lower triangular matrices with positive elements on the diagonal and SPD matrices, allowing one to transfer Euclidean structure and kernels to SPD matrices.
Though simple and natural, these Riemannian structures are often too simplistic and poorly capture the geometry of SPD matrices~\cite{Feragen_MAVA17}.

Another approach is to consider the map $\m{A} \mapsto \m{A} \m{A}^{\top}$ as a function from the space $\operatorname{GL}(d)$ of invertible matrices to $\c{S}_{\ty{++}}^d$.
This map is not one-to-one: specifically, $\m{A} \m{A}^{\top} = \m{B} \m{B}^{\top}$ if and only if $\m{A} = \m{B} \m{Q}$ where $\m{Q}$ is an orthogonal matrix: see Appendix~\ref{appdx:theory} for details.
To make this map one-to-one, we need to switch the domain to the quotient space of $\operatorname{GL}(d)$ by the equivalence relation of being equal up to an orthogonal matrix.
The resulting space $\operatorname{GL}(d) / \operatorname{O}(d)$ is a Riemannian manifold.
The Riemannian structure defined this way is not Euclidean and the resulting manifold is non-compact, hence the method of~\textcite{Borovitskiy_20NeurIPS} does not apply here either.
Still, explicit formulas for the corresponding heat kernels can be found in literature for cases of $d = 2$ and $d = 3$, which are common dimensions in robotics problems.
Specifically, for $d = 2$ there is a relatively simple formula.
Take $\m{X}, \m{Y} \in \c{S}_{\ty{++}}^2$, denote by $\m{X}_C, \m{Y}_C$ their Cholesky factors and by $H_1 \geq H_2$ the singular values of the matrix $\m{X}_C \m{Y}_C^{-1}$, and $\alpha = H_1 - H_2$, then, by \textcite{sawyer1992}, we have
\[ \label{eqn:spd_heat}
k_{\infty, \kappa, \sigma^2}(\m{X}, \m{Y})
=
\frac{\sigma^2}{C_{\infty}} \exp\del{-\frac{H_1^2 + H_2^2}{2 \kappa^2}} \int_0^{\infty} \frac{(2 s + \alpha) \exp\del{-s (s + \alpha)/ \kappa^2}}{\del{\sinh(s) \sinh(s + \alpha)}^{1/2}} \d s ,
\]
where the one-dimensional integral on the right-hand side may be evaluated numerically and $C_{\infty}$ is a normalizing constant ensuring that $k_{\infty, \kappa, \sigma^2}(\m{X}, \m{X}) = \sigma^2$.
Additional details are given in Appendix~\ref{appdx:theory}.

\paragraph{Product kernels on manifolds.}
\label{sec:product_kernel}
If the Riemannian manifold $\c{M}$ of interest is the product of $d$ manifolds $\c{M} = \c{M}_1 \x \ldots \x \c{M}_d$, where each factor $\c{M}_j$ is equipped with a Matérn kernel $k^{\c{M}_j}_{\nu, \kappa, \sigma^2}$, then we can introduce the following product Matérn kernel as a kernel on the product manifold $\c{M}$.
This is done by writing
\[
k^{\c{M}}_{\nu, \v{\kappa}, \sigma^2}(\v{x}, \v{x}')
=
\sigma^2
k^{\c{M}_1}_{\nu, \kappa_1, 1}(x_1, x_1')
\cdot
\ldots
\cdot
k^{\c{M}_d}_{\nu, \kappa_d, 1}(x_d, x_d'),
\]
where $\v{\kappa} = \del{\kappa_1, \ldots, \kappa_d}$, $\v{x} = \del{x_1, \ldots, x_d}$ and $\v{x}' = \del{x_1', \ldots, x_d'}$.
Note that this enables automatic relevance determination in the product Riemannian setting.
Because of this, it is an alternative and more expressive way to define a kernel on e.g. a torus or other product manifold.
Note that the product of Matérn kernels does not generally coincide with the Matérn kernel on the product manifold, unless $\nu = \infty$ and $\kappa_1 = .. = \kappa_d$.
The product of manifolds naturally arises when representing the pose of rigid bodies such as the end-effector of a robotic manipulator, which lies on $\R^3 \times \mathbb{S}^3$.

\vspace{-0.1cm}
\section{Experiments} \label{sec:exp}
We evaluate the performance of geometry-aware Bayesian optimization via Riemannian Matérn kernels on a set of benchmark functions and robotics settings, using expected improvement~\cite{Mockus75:EI} as the acquisition function. 
Each optimization process is repeated $30$ times with $5$ random initial samples. 
Our implementations employ GPyTorch~\cite{Gardner18:Gpytorch}, BoTorch~\cite{Balandat19:Botorch} and Pymanopt~\cite{PyManOpt16} as well as the Robotics Toolbox~\cite{RoboticsToolboxPython}. Cost parameter values and additional experiments are given in Appendix \ref{apdx:experiments}. 

\vspace{-0.1cm}
\subsection{Benchmarks}
We first study performance using benchmark test functions projected onto the manifolds $\mathbb{S}^5$, $\operatorname{SO}(3)$, $\c{S}_{\ty{++}}^2$, and $\c{H}^3$. 
The search domain $\c{X}$ is given by the full manifold, except for $\c{S}_{\ty{++}}^d$ where we restrict to matrices with eigenvalues $\lambda_i \in [0.001; 5]$.  
For each manifold, we compare geometry-aware Bayesian optimization against classical Bayesian optimization with Euclidean kernels using constrained optimization on the Euclidean space to minimize the acquisition function. 
For the sphere and SPD manifolds, we additionally compare against the scheme used by~\textcite{Jaquier19CoRLa} featuring a naïve Riemannian squared exponential kernel, where Euclidean distance is replaced by the geodesic distance. 
For the SPD manifold, we consider two Riemannian kernels, namely~\eqref{eqn:spd_heat}, and a product of kernels $\mathbb{R}^2\times \mathbb{S}^2$ which is applied on the eigendecomposition of SPD matrices. 
We also compare against an alternative implementation using the Cholesky decomposition of an SPD matrix $\m{X} = \m{L}\m{L}^\trsp$, so that the resulting parameter is the vectorization of the lower triangular matrix $\m{L}$.

\begin{figure}
\centering
\includegraphics{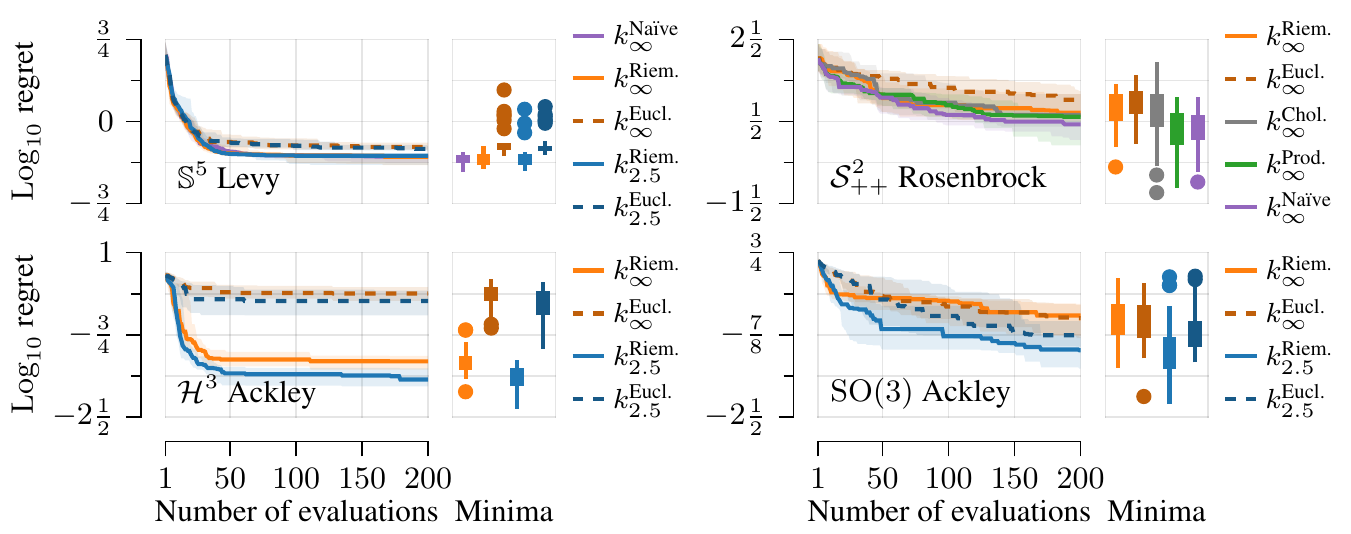}
\caption{Logarithm of the regret for a set of benchmark test functions on $\mathbb{S}^5$, $\operatorname{SO}(3)$, $\c{S}_{\ty{++}}^2$, and $\c{H}^3$.} 
\label{Fig:Benchmark}
\end{figure}

Figure~\ref{Fig:Benchmark} shows performance results, including evolution of the median and the distribution of the logarithm of the simple regret of the final recommendation $\bm{x}_N$ after $200$ iterations.
We observe that geometry-aware algorithms generally match or outperform their Euclidean counterparts.
These differences are more pronounced in higher dimensions, and for manifolds which more significantly depart from the Euclidean setting due to presence of curvature or other geometric features, with the hyperbolic manifold $\c{H}^3$ showing the most pronounced difference.
Surprisingly, the naïve Riemannian squared exponential kernel---which for certain length scales may become ill-defined---performs competitively in some settings: this may occur because small length scales keep it out of the problematic regime, and its simple form introduces less numerical error than series truncations or other approaches.
Geometry-aware Matérn kernels ($\nu=2.5$) outperform the squared exponential ones for the test functions with uneven landscapes, such as the Ackley function in Figure~\ref{Fig:Benchmark}.
Overall, we see that taking geometry into account generally results in faster, more data-efficient convergence and lower variance, at cost of more complex algorithmic implementation.

\subsection{Robotics Experiments}
\label{subsec:robotic_experiments}
Here, we evaluate the performance of geometry-aware Bayesian optimization on several simulated robotics scenarios.
In the first experiment, we use Bayesian optimization as an orientation sampler aiming at satisfying the requirements defined by a cost function, similarly to the experiment presented in~\textcite{Jaquier19CoRLa}. 
A velocity-controlled robot samples an orientation reference $\v{x} = \hat{\v{q}}$ around a prior orientation $\tilde{\v{q}}$.
The objective of the robot is to minimize the absolute value of the difference between the prior and the current end-effector orientation $\v{q}$ with low joint torques $\v{\tau}$ and a high-volume manipulability ellipsoid, i.e., $f(\v{q}) = \sum w_{\v{q}} \|\operatorname{Log}_{\v{q}}(\tilde{\v{q}})\|_1 +w_{\v{\tau}} \|\v{\tau}\|_2^2 - w_{\m{M}} \operatorname{det}(\m{M})$, where $\operatorname{det}(\m{M})$ is the determinant of the velocity manipulability ellipsoid, proportional to its volume.

Next, we consider a task-compatible manipulability optimization scenario, a common problem in robotics where most solutions do not exploit the geometry of manipulability ellipsoids~\cite{Guilamo2006,InhoLee2016}.
Here, an $8$-degree-of-freedom planar robot is required to track a desired Cartesian velocity trajectory leading to a vertical line, while tracking a desired manipulability ellipsoid in its nullspace~\cite{Jaquier21IJRR}. 
Optimization aims at finding the desired manipulability $\v{x} = \widehat{\m{M}}\in\c{S}_{\ty{++}}^2$ so that the end-effector trajectory jerk $\dddot{\v{p}}$ is minimized and the robot tracks the desired manipulabity aligning with the end-effector movement direction as proposed by~\textcite{Chiacchio90}.
This leads to $f(\widehat{\m{M}}) = \sum w_{\dddot{\v{p}}}\norm{\dddot{\v{p}}}_2 + w_{\m{M}} \|\operatorname{Log}_{\m{M}}(\widehat{\m{M}})\|_1 + w_{\v{t}} \v{t}^\trsp \m{M}\v{t}$, with $\m{M}$ the current manipulability and $\v{t}$ the unit vector tangential to the end-effector path.

Finally, we consider a simple path planning problem on $\R^2$ for a robot represented by a point. 
Our goal is to find a collision-free path between two points in the environment.
To do so, we leverage geometry-aware Bayesian optimization to estimate paths for sampling, similarly to~\textcite{Marchant2014}, but introducing hyperbolic geometry.
Specifically, we discretize the robot environment by defining a grid on $\R^2$ with $4$-connected nodes, represented by a connected undirected graph $\c{G} = (V, E)$, of $n$ vertices $V$ and edges $E$.
A circle-shaped obstacle of unknown location occupies this environment: the robot finds this by sensing collisions while testing sampled paths.
The graph $\c{G}$ is mapped to $\c{H}^2$ by learning embeddings in the Lorentz model of hyperbolic geometry, similarly to~\textcite{Nickel2018}. 
At each iteration, the algorithm samples a set of $m$ points $\{\v{x}_1, \ldots \v{x}_m\} \in \c{H}^2$ representing a hyperbolic path of $m$ nodes, which are mapped back to a grid in $\R^2$ in order for the robot to navigate the environment using the resulting path.
An optimal path corresponds to the shortest collision-free trajectory from a fixed starting location to the goal $\v{x}_g$. 
We minimize the cost $f(\v{x}_1,\ldots,\v{x}_m) = \sum w_g \operatorname{dist}^{\v{x}_g}(\v{x}_1,\ldots,\v{x}_m) + w_d \operatorname{dist}^{t}(\v{x}_1,\ldots,\v{x}_m)$, where $\operatorname{dist}^{\v{x}_g}(\cdot)$ is the distance to the goal $\v{x}_g$ (zero in no-collision cases), and $\operatorname{dist}^{t}(\cdot)$ is the total travel distance.  

\begin{figure}
\centering
\includegraphics{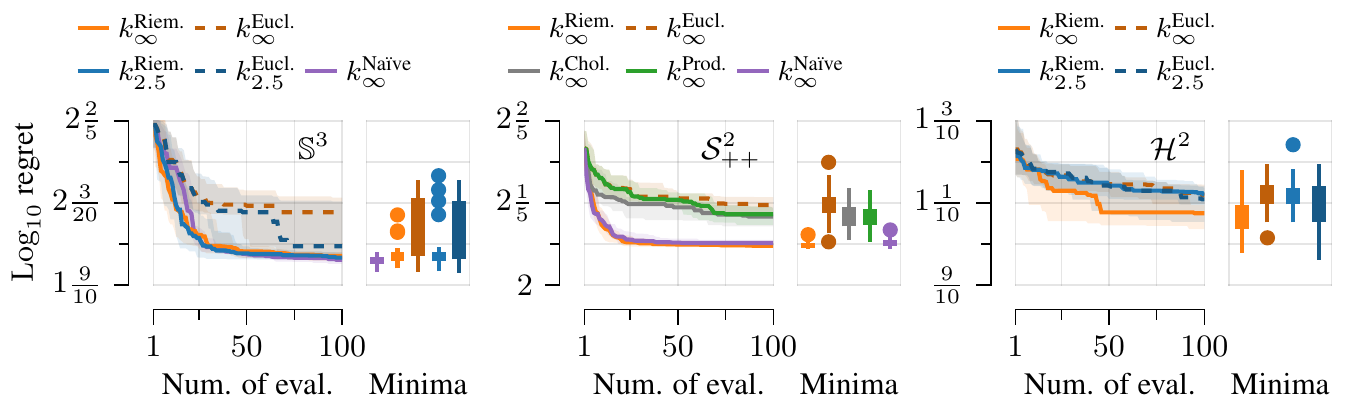}
\caption{Logarithm of the regret for the orientation, task-compatible manipulability, and path planning problems.}
\vspace{-0.4cm}
\label{Fig:RoboticsExperiments}
\end{figure}

Figure~\ref{Fig:RoboticsExperiments} shows performance results.
Here, we also see a moderate improvement in performance when using geometry-aware Bayesian optimization, mirroring the results on benchmark functions.
Note that the gained performance improvements may depend on both the manifold geometry and cost function landscape. 
For instance, the additional orientation experiment reported in Appendix~\ref{apdx:experiments} show less-pronounced
improvements as the low-dimensional nature may make the Euclidean and Riemannian kernels very similar.
This contrasts with  the results in the manipulability and path planning problems, whose geometries depart more from the Euclidean setting.

\section{Conclusion}
We study geometry-aware Bayesian optimization for robotics, and propose techniques for computing geometry-aware kernels on various Riemannian manifolds occurring in robotics applications.
This is achieved by studying how general expressions for Matérn kernels based on stochastic partial differential equations simplify for the manifolds of interest.
This provide practitioners with explicit formulas, enabling the use of geometry-aware Bayesian optimization in novel settings.
We demonstrate our techniques both on benchmarks and robotics examples to showcase improved performances and potential use cases.
We hope these contributions enable greater use of geometry among roboticists.

\acknowledgments{
NJ and TA were supported by the Carl Zeiss Foundation under the project JuBot (Jung Bleiben mit Robotern).
VB was supported by the Ministry of Science and Higher Education of the Russian Federation, agreement N\textsuperscript{\underline{o}} 075-15-2019-1620 and by "Native towns", a social investment program of PJSC Gazprom Neft.}

\printbibliography

\newpage
\appendix

\section{Experiments}
\label{apdx:experiments}
\subsection{Experimental Details}
All the robotic experiments reported in Section~\ref{subsec:robotic_experiments} used $5$ initial random samples before starting the Bayesian optimization loop.
For the orientation experiment, the parameters of the cost $f(\v{q})$ were fixed to $\tilde{\v{q}}=\begin{pmatrix}0.8 & 0.6 & 0 & 0 \end{pmatrix}^\trsp$, $w_{\v{q}}=1.0$, $w_{\v{\tau}}=1e^{-4}$, and $w_{\m{M}}=1$.
We used a velocity-based orientation controller $\dot{\v{\theta}}_r = \m{J}(\v{\theta})^\dagger\left[\dot{\v{q}}_d + \m{K}_p \operatorname{Log}_{\tilde{{\v{q}}}}(\v{q}) \right]$, with $\v{\theta}$ being the robot joint position,  $\m{J}(\v{\theta})$ representing the robot Jacobian, and $\dagger$ denoting the pseudo-inverse operator.
Furthermore, $\operatorname{Log}(\cdot)$ measures the quaternion tracking error on the sphere manifold, and $\m{K}_p = 6.0 \m{I}_4$. 

For the manipulability optimization task, the weights in the cost $f(\widehat{\m{M}})$ were set to $w_{\dddot{\v{p}}}=0.1$, $w_{\m{M}}=0.1$, and $w_{\v{t}}=15$. 
We used velocity-based control of the form 
$\dot{\v{\theta}}_r = \m{J}(\v{\theta})^\dagger\left[\dot{\v{x}}_d + \m{K}_p(\v{x}_d - \v{x})\right] + \m{N}(\v{\theta}) \c{J}(\m{M})_{(3)}^\dagger \left[\m{K}_M \text{vec}\left(\operatorname{Log}_{\widehat{\m{M}}}(\m{M})\right)\right]$, with $\v{\theta}$ and $\v{x}$ denoting the robot joint and Cartesian position, while $\m{J}(\v{\theta})$ is the robot Jacobian and $\m{N}(\v{\theta})$ is the Jacobian nullspace. 
Moreover, $\operatorname{Log}(\cdot)$ measures the manipulability tracking error on the SPD manifold, which is mapped to joint velocities via the pseudo-inverse of the manipulability Jacobian $\c{J}(\m{M})$. In addition, $\text{vec}(\cdot)$ denotes the vectorization of a matrix, and $(\cdot)_{(3)}$ denotes the 3-mode matricization of a tensor. We refer the interested reader to~\textcite{Jaquier21IJRR} for a complete description of the manipulability controller.
The position and manipulability tracking gains, for the main and secondary control tasks, correspond to $\m{K}_p = 75 \m{I}_2$ and $\m{K}_M = 7.0 \m{I}_3$. 

Finally, for the simple path planning problem, we considered a $9\times 9$ grid as a discretization of the robot planar environment, which included a circle-shaped obstacle of radius $1$ centered in $(4,4)$. 
The nodes inter-distance was $1$ unit. 
The simple graph to optimize consisted of a single branch of $5$ nodes connecting the robot start location to the goal.  
The parameters of the cost $f(\v{x}_{1,\ldots,m})$ were fixed to $w_g=1$, and $w_d=25$.

\subsection{Additional Benchmark Experiments}
\begin{figure}[t!]
\centering
\includegraphics{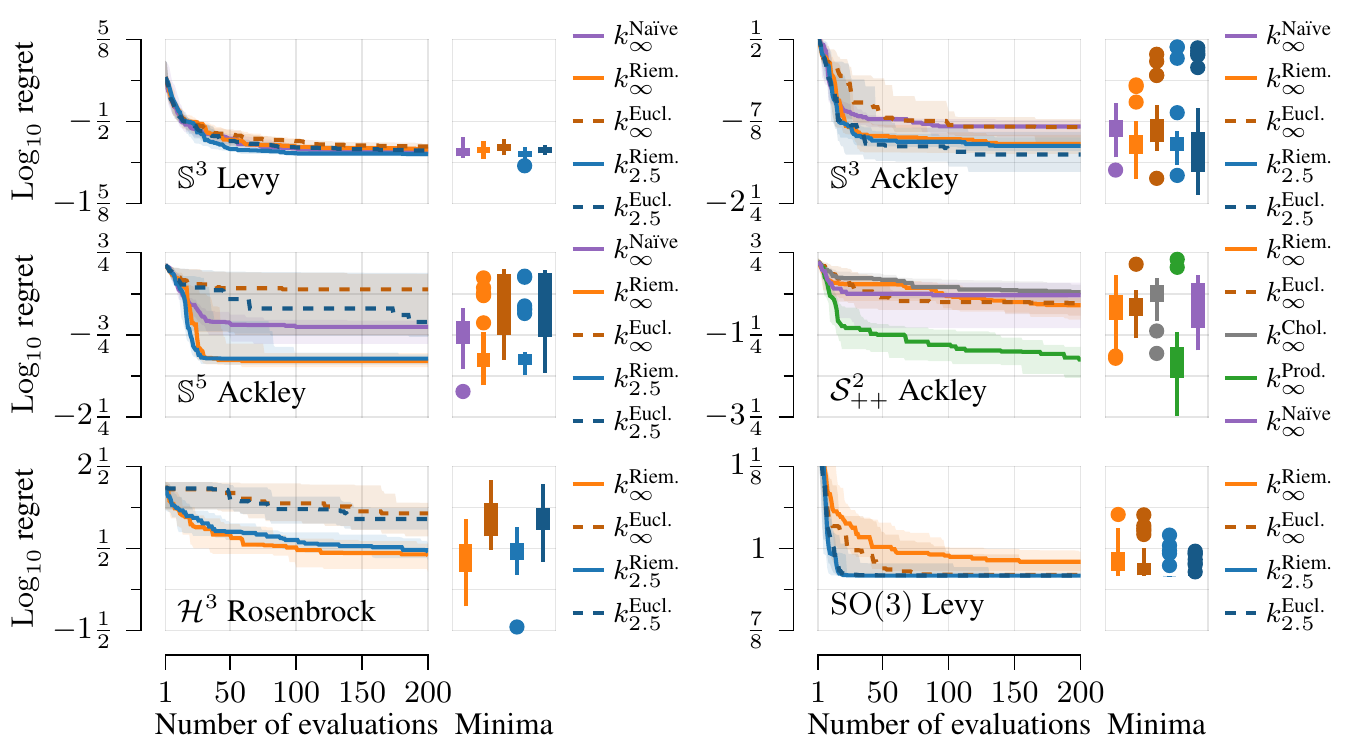}
\caption{Logarithm of the regret for a set of additional benchmark test functions on the  manifolds $\mathbb{S}^3$, $\mathbb{S}^5$, $\operatorname{SO}(3)$, $\c{S}_{\ty{++}}^2$, and $\c{H}^3$, for both geometry-aware and Euclidean Bayesian optimization algorithms.}
\label{Fig:AdditionalBenchmark}
\end{figure}

Figure~\ref{Fig:AdditionalBenchmark} shows the performance results for additional benchmark functions projected onto the manifolds $\mathbb{S}^3$, $\mathbb{S}^5$, $\operatorname{SO}(3)$, $\c{S}_{\ty{++}}^2$, and $\c{H}^3$. The left graphs show the
evolution of the median for the different Bayesian optimization approaches, and the right graphs display the distribution of the logarithm of the simple regret of the final recommendation $\bm{x}_N$ after $200$ iterations.
Similarly to the experiments presented in the main paper, we observe that geometry-aware algorithms generally match or outperform their Euclidean counterparts, resulting in better solution quality, faster convergence rate, and lower variance. 
These differences are more pronounced in higher dimensions, for instance when comparing $\mathbb{S}^5$ against to $\mathbb{S}^3$, and for manifolds which differ significantly from the Euclidean space.

\subsection{Additional Robotics Experiments}
\begin{figure}
\centering
\includegraphics{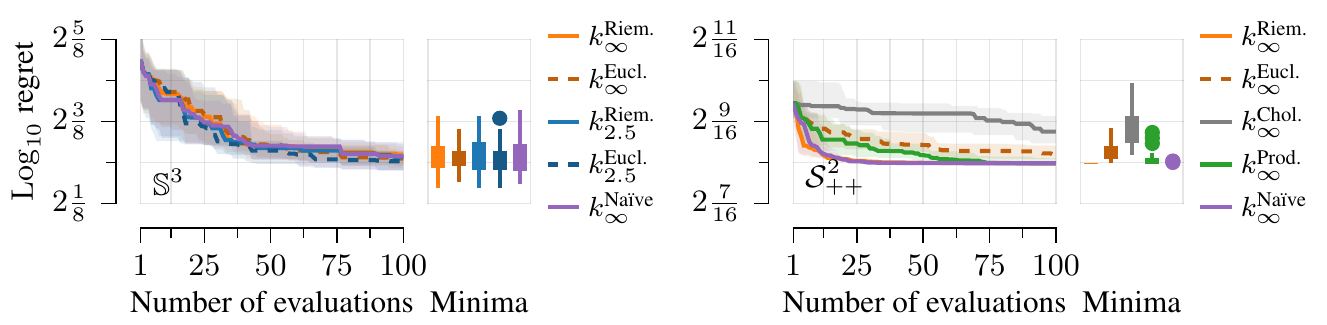}
\caption{Logarithm of the regret for the additional orientation, and task-compatible manipulability problems, for both geometry-aware and Euclidean Bayesian optimization algorithms.}
\label{Fig:AdditionalRoboticsExperiments}
\end{figure}

Figure~\ref{Fig:AdditionalRoboticsExperiments} shows the performance results for two additional robotics experiments. 
Similarly to the first experiment of Section~\ref{subsec:robotic_experiments}, we first use Bayesian optimization as an orientation sampler, where a velocity-controlled robot samples an orientation reference $\v{x} = \hat{\v{q}}$ around a prior orientation $\tilde{\v{q}}$. 
The objective of the robot is to minimize the error between the prior and the current end-effector orientation with low joint torques and an isotropic manipulability ellipsoid, i.e., $f(\v{q}) = \sum w_{\v{q}} \operatorname{dist}_{\c{S}^3}(\tilde{\v{q}}, \v{q}) +w_{\v{\tau}} \norm{\v{\tau}}_2^2 + w_{\m{M}} \text{cond}(\m{M})$, where $\v{q}$ is the current end-effector orientation, $\v{\tau}$ is the joint torques, and $\text{cond}(\m{M})$ is the condition number of the velocity manipulability ellipsoid. The parameters of the cost $f(\v{q})$ were fixed to $\tilde{\v{q}}=\begin{pmatrix}0.8 & 0.6 & 0 & 0 \end{pmatrix}^\trsp$, $w_{\v{q}}=1.0$, $w_{\v{\tau}}=0.1$, and $w_{\m{M}}=1e^{-4}$.

Next, we consider a task-compatible manipulability optimization scenario, similar to the second experiment of Section~\ref{subsec:robotic_experiments}.
Namely, a $8$-degree-of-freedom planar robot is required to track a desired Cartesian velocity trajectory leading to a vertical line, while tracking a desired manipulability ellipsoid in its nullspace.
The optimizer aims at finding the desired manipulability $\v{x} = \widehat{\m{M}}\in\c{S}_{\ty{++}}^2$ so that the end-effector acceleration $\ddot{\v{p}}$ is minimized and the robot tracks the desired manipulabity with a high precision. This leads to the cost function $f(\widehat{\m{M}}) = \sum w_{\ddot{\v{p}}}\norm{\ddot{\v{p}}}_2 + w_{\m{M}}\operatorname{dist}_{\c{S}_{\ty{++}}^2}(\m{M}, \widehat{\m{M}})$ with $\m{M}$ the current manipulability. 
Here, the weights are given the following values $w_{\ddot{\v{p}}}=0.1$ and $w_{\m{M}}=1.0$. 

As shown in Figure~\ref{Fig:AdditionalRoboticsExperiments},  geometry-aware Bayesian optimization algorithms match or outperform their Euclidean counterpart, mirroring the results presented in the main paper. The performance differences are more pronounced in the experiment presented in Section~\ref{subsec:robotic_experiments} compared to this second orientation experiment due to the uneven landscape of the cost function of the former experiment. 
The results of the manipulability experiment are similar to those presented in Section~\ref{subsec:robotic_experiments}. The naïve Riemannian squared exponential kernel---which for certain length scales may become ill-defined---performs competitively in this setting. 
This may occur because of the low dimensionality of the problem, and because small length scales keep it out of problematic situations.

\section{Theory} \label{appdx:theory}

\subsection{Matérn kernels on general Riemannian manifolds and their connection to heat kernels} \label{appdx:theory:general_manifolds}

The theoretical framework defining appropriate notions of stochastic partial differential equations~(SPDEs) and Matérn kernels as kernels of the unique solutions of these equations was presented in Appendix D of \textcite{Borovitskiy_20NeurIPS} for the case of compact Riemannian manifolds.
In the non-compact case, the technique that yields explicit solutions of the SPDEs as series depending on Laplace--Beltrami eigenpairs does not apply, but the implicit definition of Matérn Gaussian processes and the corresponding kernel remain valid under mild regularity conditions.

Specifically, if we restrict our attention to connected complete Riemannian manifolds of bounded geometry, which is indeed a rather nonrestrictive assumption, then one can generalize both the functional calculus formalism and the definitions of the appropriate function spaces needed to make the SPDEs rigorously defined. 
This can be done by means of general spectral theory, which is reviewed by \textcite{devito2020}.
The theory of \textcite{Borovitskiy_20NeurIPS} in this general setting shows that the Riemannian Matérn and squared exponential kernels are the reproducing kernels of Sobolev and diffusion spaces studied by \textcite{devito2020}.
Because of this, Theorem 8 from the latter work directly implies that the Riemannian squared exponential kernel as defined by (the generalized theory of) \textcite{Borovitskiy_20NeurIPS} coincides with the heat kernels as fundamental solutions of the heat (diffusion) equation on the Riemannian manifold.
This is a crucial point for this work, since it allows us to connect to the existing literature on heat kernels in some interesting non-compact cases.

\subsection{Matérn kernels as integrals of heat kernels} \label{appdx:theory:integrals}

By Appendix~\ref{appdx:theory:general_manifolds}, Riemannian squared exponential and Matérn kernels are well defined on sufficiently regular non-compact manifolds, and Riemannian squared exponential kernels coincide with heat kernels, which are widely studied in the mathematics literature even in non-compact cases.
Because of this, it is natural to seek a way to express Riemannian Matérn kernels in terms of the corresponding heat kernels.

In the Euclidean case, this is given by equation~\eqref{eqn:matern_integral_formula}, which we prove in Appendix~\ref{appdx:theory:integrals:euclidean}.
A similar but slightly different relation holds in the compact Riemannian case, which we prove in Appendix~\ref{appdx:theory:integrals:riemannian}.
For the general Riemannian case we adopt~\eqref{eqn:matern_integral_formula} as the \emph{definition} of Matérn kernels.
We prove that the kernels defined this way are positive definite in Appendix~\ref{appdx:theory:integrals:pd}.

\subsubsection{Euclidean Matérn kernels as integral of heat kernels} \label{appdx:theory:integrals:euclidean}

The Euclidean Matérn and squared exponential kernels are given by the following formulas \cite{rasmussen2006}:
\[
&k_{\nu, \kappa, \sigma^2}(\v{x}, \v{x}')
=
\sigma^2
\frac{2^{1-\nu}}{\Gamma(\nu)}
\del[1]{\sqrt{2 \nu} \rho / \kappa}^{\nu}
K_{\nu}\del{\sqrt{2 \nu} \rho / \kappa}
,
&
&k_{\infty, \kappa, \sigma^2}(\v{x}, \v{x}')
=
\sigma^2
e^{- \frac{\rho^2}{2 \kappa^2}}
,
\]
where $\rho = \norm{\v{x}-\v{x}'}$ denotes the distance between a pair of inputs.
These kernels may be represented as inverse Fourier transforms of the corresponding spectral densities $S_{\nu, \kappa, \sigma^2}$ \cite{rasmussen2006}, that is
\[ \label{eqn:spectral_density_representation}
k_{\nu, \kappa, \sigma^2}(\v{x}, \v{x}')
=
\int_{\R^d}
S_{\nu, \kappa, \sigma^2}(\norm{\v{\xi}})
e^{2 \pi i \innerprod{\v{x}-\v{x}'}{\v{\xi}}}
\d \v{\xi}.
\]
This covers both cases, whether $\nu$ is finite or infinite.
The respective spectral densities are given by
\[
\label{eqn:matern_sd}
S_{\nu, \kappa, \sigma^2}(\lambda)
&=
\sigma^2
\overbracket[0.1ex]{\frac{2^d \pi^{d/2} \Gamma(\nu+d/2) \del{2 \nu}^{\nu}}{\Gamma(\nu) \kappa^{2 \nu}}}^{\text{denote by } 1/C_{\nu}}
\del{\frac{2 \nu}{\kappa^2} + 4 \pi^2 \lambda^2}^{-\nu-\frac{d}{2}},
\\
\label{eqn:se_sd}
S_{\infty, \kappa, \sigma^2}(\lambda)
&=
\sigma^2
\underbracket[0.1ex]{(2 \pi \kappa^2)^{d/2}}_{\text{denote by } 1/C_{\infty}}
e^{- 2 \pi^2 \kappa^2 \lambda^2}.
\]
Here $d$ is the dimension of the Euclidean space under consideration and $C_{\nu}, C_{\infty}$ are the normalizing constants that ensure $k_{\nu, \kappa, \sigma^2}(\v{x}, \v{x}) = \sigma^2$.
Note that, despite this notation, they depend on both $\nu$ and~$\kappa$.

We start verifying equation~\eqref{eqn:matern_integral_formula} by noting that by \textcite[Section 3.326, Item 2]{gradshteyn2014} we have
\[ \label{eqn:exponential_integral}
\int_0^{\infty} u^n e^{- a u} \d u = \Gamma(n+1) a^{-n-1}
.
\]
Substituting $n = \nu + d/2 - 1$ and $a = 2 \nu / \kappa^2 + 4 \pi^2 \lambda^2$ into this equation and then performing a simple rearrangement of terms, we get the following expression for $C_{\nu} / \sigma^2 S_{\nu, \kappa, \sigma^2}(\lambda)$:
\[
\del{\frac{2 \nu}{\kappa^2} + 4 \pi^2 \lambda^2}^{-\nu - \frac{d}{2}}
&=
\Gamma(\nu + d/2)^{-1}
\int_0^{\infty}
u^{\nu + \frac{d}{2}-1}
e^{-\frac{2 \nu}{\kappa^2} u}
e^{-4 \pi^2 \lambda^2 u}
\d u,
\\
&=
\del{4 \pi}^{-\frac{d}{2}}\Gamma(\nu + d/2)^{-1}
\int_0^{\infty}
u^{\nu - 1}
e^{-\frac{2 \nu}{\kappa^2} u}
(4 \pi u)^{\frac{d}{2}} e^{-4 \pi^2 \lambda^2 u}
\d u,
\\ \label{eqn:aux_sd_formula_eucl}
&=
\sigma^{-2} \del{4 \pi}^{-\frac{d}{2}}\Gamma(\nu + d/2)^{-1}
\int_0^{\infty}
u^{\nu - 1}
e^{-\frac{2 \nu}{\kappa^2} u}
S_{\infty, \sqrt{2 u}, \sigma^2}(\lambda)
\d u
.
\]

Now, using equations~\eqref{eqn:spectral_density_representation}~and~\eqref{eqn:matern_sd} and then ~\eqref{eqn:aux_sd_formula_eucl}, we write
\[
k_{\nu, \kappa, \sigma^2}(\v{x}, \v{x}')
=
\frac{\sigma^2}{C_{\nu}}
&\int_{\R^d}
\del{\frac{2 \nu}{\kappa^2} + 4 \pi^2 \norm{\v{\xi}}^2}^{-\nu - \frac{d}{2}}
e^{2 \pi i \innerprod{\v{x}-\v{x}'}{\v{\xi}}}
\d \v{\xi}
\\
=
\frac{1}{C_{\nu} \del{4 \pi}^{\frac{d}{2}}\Gamma(\nu + d/2)}
&\int_{\R^d}
\int_0^{\infty}
u^{\nu - 1}
e^{-\frac{2 \nu}{\kappa^2} u}
S_{\infty, \sqrt{2 u}, \sigma^2}(\norm{\v{\xi}})
\d u
\,
e^{2 \pi i \innerprod{\v{x}-\v{x}'}{\v{\xi}}}
d \v{\xi} = \ldots
\]
By changing the order of integration, rearranging terms and using formulas~\eqref{eqn:spectral_density_representation}~and~\eqref{eqn:se_sd} we get
\[ \label{eqn:matern_integral_formula_full}
\ldots =
\frac{1}{C_{\nu} \del{4 \pi}^{\frac{d}{2}}\Gamma(\nu + d/2)}
\int_0^{\infty}
u^{\nu - 1}
e^{-\frac{2 \nu}{\kappa^2} u}
k_{\infty, \sqrt{2 u}, \sigma^2}(\v{x}, \v{x}')
\d u,
\]
which coincides with \eqref{eqn:matern_integral_formula} up to a multiplicative constant which we disregard for convenience since it only affects the normalization of the kernel.

\subsubsection{Compact Riemannian Matérn kernels as integral of heat kernels} \label{appdx:theory:integrals:riemannian}

Normalizing constants in the definition of the Riemannian Matérn and squared exponential kernels are not available in closed form and, at the same time, depend on the length scale.
To get around this, we connect the unnormalized versions of compact Riemannian Matérn kernels to their respective unnormalized squared exponential kernels.
As a result, the relation turns out to be slightly different, but still very similar to~\eqref{eqn:matern_integral_formula}.

First, we introduce the notation similar to \eqref{eqn:matern_sd} and \eqref{eqn:se_sd}, where the superscript $u$ stands for \emph{unnormalized}:
\[
\label{eqn:sd_riemannian}
&S^{u}_{\nu, \kappa}(\lambda)
=
\del{\frac{2 \nu}{\kappa^2} + \lambda}^{-\nu-\frac{d}{2}}
&
&S^{u}_{\infty, \kappa}(\lambda)
=
e^{- \frac{\kappa^2}{2} \lambda}.
\]
Then for both finite and infinite $\nu$ we have $k_{\nu, \kappa, \sigma^2}(x, x') = \sigma^2 / C_{\nu} \sum_{n=0}^{\infty} S_{\nu, \kappa}^{u}(\lambda_n) f_n(x)f_n(x')$.
We~are going to relate the unnormalized kernels given, both for finite and infinite $\nu$, by
\[
k^{u}_{\nu, \kappa}(x, x') = \sum_{n=0}^{\infty} S_{\nu, \kappa}^{u}(\lambda_n) f_n(x)f_n(x').
\]

As in the Euclidean case we use ~\eqref{eqn:exponential_integral}, but this time with $n = \nu + d/2 - 1$ and $a = 2 \nu / \kappa^2 + \lambda$, where $n$ stays as before and $a$ is slightly different, and then perform calculations similar to~\eqref{eqn:aux_sd_formula_eucl}, obtaining
\[
S_{\nu, \kappa}^{u}(\lambda)
=
\del{\frac{2 \nu}{\kappa^2} + \lambda}^{-\nu - \frac{d}{2}}
=
\Gamma(\nu + d/2)^{-1}
&\int_0^{\infty}
u^{\nu + \frac{d}{2}-1}
e^{-\frac{2 \nu}{\kappa^2} u}
e^{-\lambda u}
\d u
\\ \label{eqn:aux_sd_formula_mani}
=
\Gamma(\nu + d/2)^{-1}
&\int_0^{\infty}
u^{\nu +\frac{d}{2} - 1}
e^{-\frac{2 \nu}{\kappa^2} u}
S_{\infty, \sqrt{2 u}}^{u}(\lambda)
\d u
.
\]
Now, write for finite $\nu$
\[
k^{u}_{\nu, \kappa}(x, x')
&=
\sum_{n=0}^{\infty} S_{\nu, \kappa}^{u}(\lambda_n) f_n(x)f_n(x')
\\
&=
\sum_{n=0}^{\infty}
\Gamma(\nu + d/2)^{-1}
\int_0^{\infty}
u^{\nu +\frac{d}{2} - 1}
e^{-\frac{2 \nu}{\kappa^2} u}
S_{\infty, \sqrt{2 u}}^{u}(\lambda)
\d u
f_n(x)f_n(x')
\\
&=
\Gamma(\nu + d/2)^{-1}
\int_0^{\infty}
u^{\nu +\frac{d}{2} - 1}
e^{-\frac{2 \nu}{\kappa^2} u}
\sum_{n=0}^{\infty}
S_{\infty, \sqrt{2 u}}^{u}(\lambda)
f_n(x)f_n(x')
\d u
\\
&=
\Gamma(\nu + d/2)^{-1}
\int_0^{\infty}
u^{\nu +\frac{d}{2} - 1}
e^{-\frac{2 \nu}{\kappa^2} u}
k_{\infty, \sqrt{2 u}}^{u}(x, x')
\d u
\]
which yields the expression
\[ \label{eqn:matern_integral_formula_comp}
k^{u}_{\nu, \kappa}(x, x')
=
\Gamma(\nu + d/2)^{-1}
\int_0^{\infty}
u^{\nu +\frac{d}{2} - 1}
e^{-\frac{2 \nu}{\kappa^2} u}
k_{\infty, \sqrt{2 u}}^{u}(x, x')
\d u
\]
which is of the form we were looking for.

\subsubsection{Integral Matérn kernels are positive (semi)definite whenever heat kernels are} \label{appdx:theory:integrals:pd}

Assume $k_{\nu, \kappa, \sigma^2}$ is defined by~\eqref{eqn:matern_integral_formula} and that the corresponding heat kernel $k_{\infty, \kappa, \sigma^2}$ is positive definite.\footnote{Heat kernels on connected complete Riemannian manifolds of bounded geometry are positive definite, this follows, for example, from the discussion in Appendix~\ref{appdx:theory:general_manifolds}. Their approximations used for practical Gaussian process regression are typically positive semi-definite.}
Take some locations $x_1, \ldots, x_n = \v{x}$ and consider the matrix $\m{K}^{\nu}_{\v{x} \v{x}}$ with elements $k_{\nu, \kappa, \sigma^2}(x_i, x_j)$.
In order to prove that $k_{\nu, \kappa, \sigma^2}$ is positive definite we need to show that $\m{K}^{\nu}_{\v{x} \v{x}}$ is a positive definite matrix for an arbitrary choice of $n$ and $\v{x}$.
This means that we need to show that $\v{y}^{\top} \m{K}^{\nu}_{\v{x} \v{x}} \v{y}^{\top} > 0$ for all nonzero vectors $\v{y} \in \R^n$.

To prove that $\m{K}^{\nu}_{\v{x} \v{x}}$ is positive definite, consider the matrices $\m{K}^{\infty, \kappa}_{\v{x} \v{x}}$ with elements $k_{\infty, \kappa, \sigma^2}(x_i, x_j)$.
Then, extending equation~\eqref{eqn:matern_integral_formula} to matrices, we have
\[
\m{K}^{\nu}_{\v{x} \v{x}}
=
\int_0^{\infty}
  u^{\nu - 1}
  e^{-\frac{2 \nu}{\kappa^2} u}
  K^{\infty, \sqrt{2 u}}_{\v{x} \v{x}}
  \d u.
\]
Because of this, we obtain
\[
\v{y}^{\top} \m{K}^{\nu}_{\v{x} \v{x}} \v{y}
=
\int_0^{\infty}
  \underbracket[0.1ex]{
  u^{\nu - 1}
  e^{-\frac{2 \nu}{\kappa^2} u}
  \vphantom{\v{y}^{\top} K^{\infty, \sqrt{2 u}}_{\v{x} \v{x}} \v{y}}
  }_{*}
  \underbracket[0.1ex]{\v{y}^{\top} K^{\infty, \sqrt{2 u}}_{\v{x} \v{x}} \v{y}}_{**}
  \d u,
\]
where the factor $*$ of the integrand is obviously positive and the factor $**$ of the integrand is positive because $\m{K}^{\infty, \sqrt{2 u}}_{\v{x} \v{x}}$ is positive definite by assumption, thus the integral is positive.
Thus $k_{\nu, \kappa, \sigma^2}$ is positive definite.
The argument transfers to the case of positive semi-definiteness mutatis mutandis.

\subsection{Matérn kernels for the torus}
On the torus $\T^d$ (with flat metric), every vector of integers $\v{\tau} \in \Z^d$ corresponds to the eigenvalue $\lambda_{\v{\tau}} = 4 \pi^2 \norm{\tau}^2$ and to the two orthonormal eigenfunctions $f_{\v{\tau}, 1}(\v{x}) = \sqrt{2}\cos(2\pi \innerprod{\v{\tau}}{\v{x}})$ and $f_{\v{\tau}, 2}(\v{x}) = \sqrt{2}\sin(2\pi \innerprod{\v{\tau}}{\v{x}})$, unless of course $\v{\tau} = 0$, which corresponds to the eigenvalue $\lambda_{\v{0}} = 0$ and a single eigenfunction $f_{\v{\tau}}(\v{x}) = 1$: for details, see \textcite{gordon2000}.
Note that the eigenvalues corresponding to distinct $\v{\tau}, \v{\tau}' \in \Z^d$ may coincide, take e.g. $\v{\tau} = (15, 20)$ and $\v{\tau'} = (7, 24)$.
However, the eigenfunctions corresponding to distinct $\v{\tau}, \v{\tau}' \in \Z^d$ will always be different (and orthogonal) unless $\v{\tau} = - \v{\tau}$.
Finally, observe that
\[ \label{eqn:torus_geg}
f_{\v{\tau}, 1}(\v{x}) f_{\v{\tau}, 1}(\v{x}')
+
f_{\v{\tau}, 2}(\v{x}) f_{\v{\tau}, 2}(\v{x}')
=
2 \cos(2 \pi \innerprod{\v{\tau}}{\v{x} - \v{x}'})
\]
for $\v{\tau} \not= 0$ because of the identity $\cos(x-y) = \cos(x) \cos(y) + \sin(x) \sin(y)$.

Since the eigenpairs are parameterized by vectors of integers instead of just natural numbers, it is thus convenient to reorder the series in \eqref{eqn:mani-matern-formula} accordingly---this can be done because the series converges unconditionally \cite{Borovitskiy_20NeurIPS}.
Then, the summation should only be performed over \emph{half} of $\Z^d$ to exclude the possibility of accounting non-orthogonal (even linearly dependent) eigenfunctions twice: for $\v{\tau} \in \Z^d$, $\v{\tau} \not= 0$ and for $-\v{\tau}$.
Instead of changing the summation index set, we notice that the expression on the right-hand side of \eqref{eqn:torus_geg} does not change when $\v{\tau}$ is substituted for $-\v{\tau}$ because cosine is an even function and divide the summands for $\v{\tau} \not= 0$ by two, turning $2 \cos(2 \pi \innerprod{\v{\tau}}{\v{x} - \v{x}'})$ into $\cos(2 \pi \innerprod{\v{\tau}}{\v{x} - \v{x}'})$.
This leads to the expression of the form given by \eqref{eqn:heat_torus} for the Matérn and squared exponential kernels, namely
\[
k_{\nu, \kappa, \sigma^2}(\v{x},\v{x}') = \frac{\sigma^2}{C_{\nu}}
\sum_{\v{\tau} \in \Z^d}
\Phi(4 \pi^2 \norm{\tau}^2)
\cos(2 \pi \innerprod{\v{\tau}}{\v{x}-\v{x}'}),
\]
where $\Phi$ is as in \eqref{eqn:mani-matern-formula}.
Note that the case of $\v{\tau} = \v{0}$ is properly accounted for because $\cos(2 \pi \innerprod{\v{0}}{\v{x}-\v{x}'}) = 1$.

\subsection{Heat and Matérn kernel for the special orthogonal group}

The formula for the heat (squared exponential) kernel for the special orthogonal group $\operatorname{SO}(d)$ is available in the literature: see e.g. \textcite[page 12]{wong2006}, and \textcite[page 45]{stein1970}.
As described in the main text, the expression for the kernel is
\[
k_{\infty, \kappa, \sigma^2}(\m{X}, \m{Y})
=
\frac{\sigma^2}{C_{\infty}}\sum_{\pi}
e^{-\frac{\kappa^2}{2} \lambda_{\pi}} d_{\pi} \chi_{\pi}(\m{X} \m{Y}^{-1}).
\tag{\ref{eqn:sod_heat}}
\]
We now discuss the precise meaning of the symbols $\pi$, $\lambda_{\pi}$, $d_{\pi}$, $\chi_{\pi}$ in equation~\eqref{eqn:sod_heat} and how to compute their values in practice.

First, the summation variable $\pi$ runs over the set of \emph{highest weights}---these are effectively tuples of non-negative integers that enumerate \emph{irreducible representations} of the group, i.e. homomorphisms $\pi\colon\operatorname{SO}(d)\to\operatorname{GL}(V)$ for some (complex) vector space $V$ such that $V$ cannot be split into a direct sum of non-trivial $\pi(\operatorname{SO}(d))$-invariant subspaces.
Iterating over the highest weights amounts to iterating over the tuples of non-negative integers and then post-processing them according to the theory from \textcite[Chapter~VI, Section~2]{brockerdieck1985}.
To calculate the kernel to a given numerical precision, one can grade all such tuples by the sum of their entries, and only leave those terms of~\eqref{eqn:sod_heat} which correspond to the entries with the smallest sum.

To each \emph{highest weight} $\pi$ there corresponds a particular eigenvalue $\lambda_\pi$ of the Laplace--Beltrami operator.
It can be computed as the value of a certain quadratic polynomial in the entries of $\pi$ via the identification of Laplace--Beltrami and Casimir operators and Freudenthal's formula for the eigenvalues, which is a particular case of Freudenthal's multiplicity formula, see~\textcite[Section~22]{humphreys1994}. 
Note that eigenvalues corresponding to distinct representations can coincide.

The function $\chi_{\pi}(\m{A}) = \operatorname{tr}(\pi(\m{A}))$ is called the \emph{character} of the representation corresponding to the highest weight $\pi$.
Its values can be calculated by means of the Weyl character formula as the ratio of two explicitly known polynomials depending only on the eigenvalues of $\m{A}$. The calculation of these polynomials involves a fair bit of Lie-theoretic machinery, see \textcite[Chapter~VI, Section~1]{brockerdieck1985}, or \textcite[Section~24]{humphreys1994} for details.

Finally, the value $d_{\pi} = \chi_{\pi}(\m{I})$ evaluates to the trace of the identity matrix and is thus the dimension of the representation $\pi$. Moreover, by the Peter--Weyl Theorem---see \textcite[Chapter~III]{brockerdieck1985}---each irreducible representation $\pi$ (of dimension $n$) corresponds to an $n^2$-dimensional eigenspace for $\lambda_\pi$.

Thanks to the integral representation of Matérn kernels given by \eqref{eqn:matern_integral_formula_comp}, we may obtain an expression for them without diving into details on how the simplified expression \eqref{eqn:sod_heat} was actually obtained in prior works.
Apart from the normalizing constant, the only part of \eqref{eqn:sod_heat} that depends on the length scale, which is the variable with respect to which the integration is performed in \eqref{eqn:matern_integral_formula_comp}, is the coefficient $e^{-\frac{\kappa^2}{2} \lambda_{\pi}}$.
Swapping the order of integration and summation, we readily obtain the following formula for Matérn kernels:
\[
k_{\nu, \kappa, \sigma^2}(\m{X}, \m{Y})
=
\frac{\sigma^2}{C_{\nu}}\sum_{\pi}
\del{\frac{2 \nu}{\kappa^2} + \lambda_{\pi}}^{-\nu - \frac{d}{2}}
d_{\pi} \chi_{\pi}(\m{X} \m{Y}^{-1}).
\]

\subsection{SPD manifold as $\operatorname{GL}(d) / \operatorname{O}(d)$ and the corresponding heat kernel}

First, we prove that the map $T: \operatorname{GL}(d) / \operatorname{O}(d) \to \c{S}_{\ty{++}}^d$ such that $T \del{\m{A} \cdot \operatorname{O}(d)} = \m{A} \m{A}^T$ is well defined, i.e. does not depend on the choice of the representative $\m{A}$ of coset $\m{A} \cdot \operatorname{O}(d)$, and is bijective (and in particular, one-to-one).
The mapping $\operatorname{GL(d)}\to\c{S}_{\ty{++}}^d$ acting as $\m{A} \mapsto \m{A}\m{A}^\top$ is surjective because of the Cholesky decomposition.
On the other hand, because of the QR decomposition and uniqueness of the Cholesky decomposition, $\m{A}\m{A}^\top = \m{B}\m{B}^\top$ if and only if $\m{A}\m{B}^{-1}$ is an orthogonal matrix.
This means that $T$ is correctly defined and is bijective.
The identification of $\operatorname{GL}(d) / \operatorname{O}(d) $ and $\c{S}_{\ty{++}}^d$ through operator $T$ induces \cite{herz1978} the affine-invariant \cite{thanwerdas2019} metric over $\c{S}_{\ty{++}}^d$ with geodesic distance given by
\[
\operatorname{dist}(\m{X},\m{Y}) = \norm{\log(\m{X}^{-1/2}\m{Y}\m{X}^{-1/2})}
, 
\]
where $\norm{\m{Z}} = \sqrt{\operatorname{tr}(\m{Z}\m{Z}^\top)}$ is the Frobenius matrix norm.

For this particular choice of the metric the explicit formulas for the heat kernel in cases $d=2$ and $d=3$ are given by \textcite{sawyer1992} with the more detailed exposition given in the thesis of \textcite{sawyer1989}.
There, after observing that the kernel $k(\m{A}, \m{B})$ only depends on the matrix $\m{A}_C \m{B}_C^{-1}$ with $\m{A}_C, \m{B}_C$ the Cholesky factors of $\m{A}, \m{B}$, the explicit formulas for the one-parameter function are inferred by a particular technical argument given in that work.
In the case $d = 2$, this yields the formula \eqref{eqn:spd_heat} presented in the main text.
In the case $d = 3$, an explicit formula is provided as well, although, due to its cumbersome nature, we do not carry it over to this paper.

\end{document}